\documentclass[10pt, a4paper,conference]{IEEEtran}
\IEEEoverridecommandlockouts
\usepackage{cite}
\usepackage{amsmath,amssymb,amsfonts}
\usepackage{bm} 
\usepackage{algorithmic}
\usepackage{graphicx}
\usepackage{textcomp}
\usepackage{xcolor}
\usepackage{hyperref}
\def\BibTeX{{\rm B\kern-.05em{\sc i\kern-.025em b}\kern-.08em
    T\kern-.1667em\lower.7ex\hbox{E}\kern-.125emX}}
\begin{document}

\title{6D Pose Estimation with Correlation Fusion\\

}

\author{
\normalsize{
Yi~Cheng$^{1}$\thanks{Hongyuan Zhu is the corresponding author. Email: zhuh@i2r.a-star.edu.sg.}, Hongyuan~Zhu$^{1}$, Ying Sun$^{1}$, Cihan~Acar$^{2}$, Wei~Jing$^{2}$, Yan~Wu$^{2}$, Liyuan~Li$^{1}$, Cheston~Tan$^{1}$, Joo-Hwee~Lim$^{1}$
} \\

    \normalsize\textit{$^{1}$ Visual Intelligence} ~~~~~~~ \normalsize\textit{$^{2}$ Robotics \& Autonomous Systems} \\
    \normalsize\textit{Institute for Infocomm Research, Agency for Science, Technology and Research} \\
	\normalsize{\{cheng\_yi, zhuh, suny, acar\_cihan, jing\_wei, wuy, lyli, cheston-tan, joohwee\}@i2r.a-star.edu.sg}
	
}

\maketitle

\begin{abstract}

6D object pose estimation is widely applied in robotic tasks such as grasping and manipulation. Prior methods using RGB-only images are vulnerable to heavy occlusion and poor illumination, so it is important to complement them with depth information. However, existing methods using RGB-D data cannot adequately exploit consistent and complementary information between RGB and depth modalities. In this paper, we present a novel method to effectively consider the correlation within and across both modalities with attention mechanism to learn discriminative and compact multi-modal features. Then, effective fusion strategies for intra- and inter-correlation modules are explored to ensure efficient information flow between RGB and depth. To our best knowledge, this is the first work to explore effective intra- and inter-modality fusion in 6D pose estimation. The experimental results show that our method can achieve the state-of-the-art performance on LineMOD and YCB-Video dataset. We also demonstrate that the proposed method can benefit a real-world robot grasping task by providing accurate object pose estimation.

\end{abstract}


\section{Introduction}

6D pose estimation, which aims to predict the 3D rotation and translation from object space to camera space, is useful in 3D object detection and recognition \cite{LoghmaniPCV19, MousavianAFK17}, robot grasping and manipulation\cite{tremblay2018corl,li2018fast}. However, it remains challenging as both accuracy and efficiency are required in real-world applications.

Existing methods can be divided into RGB-only methods and RGB-D based methods. Methods with RGB-only images as input use deep neural networks to either regress 6D pose directly \cite{xiang2017posecnn,li2018deepim,manhardt2018eccv} or detect the 2D projections of 3D key points and then obtain 6D pose by solving a Perspective-n-Point(PnP) problem \cite{pavlakos2017icra,oberweger2018eccv,peng2019pvnet,hu2019cvpr,tekin2018cvpr,tremblay2018corl,rad2017bb8}. Although these methods can achieve fast inference and address certain occlusions, they still generally underperform RGB-D based methods, as the depth map can provide effective complementary information of the object geometrics \cite{zhu2016cvpr}. Most recent RGB-D based methods predict coarse 6D pose, and then use the depth map to refine the previous estimation with iterative-closest point (ICP) algorithm \cite{jafari2018ipose,kehl2017ssd,brachmann2014eccv,krull2015iccv,michel2017iccv,zeng2017icra}. However, ICP is computationally expensive and sensitive to initialization.  

\begin{figure}[htb]
	\centering
	\includegraphics[width=0.95\linewidth]{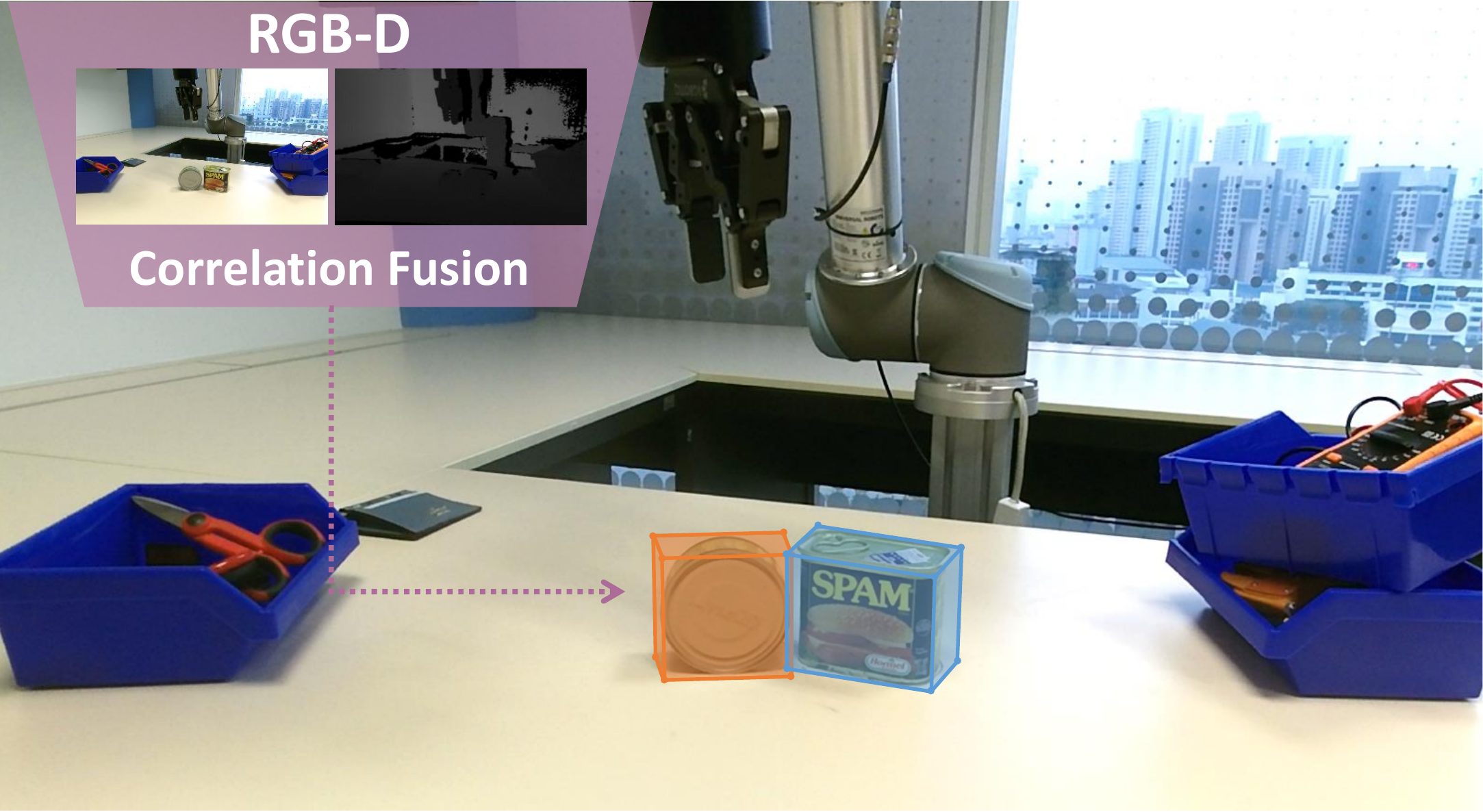}
	\caption{We develop an end-to-end deep network model for 6D pose estimation which performs effective fusion for RGB and depth features with correlation learning for fast and accurate predictions for real-time applications such as robot grasping and manipulation.}
	\label{fig:titlefigure}\vspace{-10pt}
\end{figure}

To overcome these problems, DenseFusion \cite{wang2019densefusion} proposes an RGB-D based deep neural network by simultaneously considering the visual appearance and geometry structure simultaneously. It is robust to occlusion and achieves real-time inference speed. However, this method does not consider the correlation within and between the RGB and depth modalities to fully exploit the consistent and complementary information from them to learn discriminative features for object pose estimation. 

In this paper, we propose a novel Correlation Fusion (CF) framework which models the feature correlation within and between RGB and depth modalities to improve the performance of 6D pose estimation. We propose two modules namely Intra-modality Correlation Modeling and Inter-modality Correlation Modeling, to help select prominent features within and cross two modalities using a self-attention mechanism. Furthermore, multiple strategies for fusing the intra- and inter-modality information are explored to ensure efficient and effective information flow within and between the modalities. Our experiments show that pose estimation accuracy can be further improved with the proposed fusion strategies. 

The main contributions of our work can be summarized as follows. Firstly, we propose intra- and inter-correlation modules to exploit the consistent and complementary information within and between RGB and depth modalities for 6D pose estimation. Secondly, we explore multiple strategies for fusing the intra- and inter-modality information flow to learn discriminative multi-modal features. Thirdly, we demonstrate that the proposed method can achieve the state-of-the-art performance on widely-used benchmark datasets for 6D pose estimation, including LineMOD \cite{hinterstoisser2011iccv} and YCB-Video \cite{xiang2017posecnn} datasets. Lastly, we showcase that our method can benefit robot grasping tasks by providing an accurate estimation of object pose.


\section{Related works}
In this section, we briefly review the recent works based on machine learning or deep learning for 3D object detection and pose estimation. 

\subsection{Pose Estimation from RGB Images}
6D object pose estimation from a single RGB image has been well studied in recent years. Existing methods either perform regression from detection \cite{xiang2017posecnn} or predict 2D  projection of predefined 3D key points \cite{pavlakos2017icra,oberweger2018eccv,peng2019pvnet,hu2019cvpr,tekin2018cvpr,tremblay2018corl,rad2017bb8}. The former can handle low-texture and partially occluded objects. However, the regression results are sensitive to small errors due to large search space. On the other hand, the keypoint-based methods can alleviate the issue of occlusion, but suffer from truncated objects as some of the key points may be outside the input image. Moreover, the aforementioned methods do not utilize the depth information, and therefore may not be able to disambiguate the objects' scales due to perspective projection. Compared with these methods, the proposed method is formulated to effectively fuse RGB and depth information for more accurate 6D pose estimation.

\subsection{Pose Estimation from RGB-D Images}
The performance of 6D pose estimation can be further improved by incorporating depth information. Current RGB-D based approaches utilize depth information mainly in three ways. First, RGB and depth information are used at separate stages \cite{jafari2018ipose,xiang2017posecnn,kehl2017ssd}, where a coarse 6D pose is predicted from an RGB image, followed by ICP algorithm using depth information for refinement. Second, RGB and depth modalities are fused at early stages \cite{brachmann2014eccv,krull2015iccv,michel2017iccv}, where the depth map is treated as another channel and concatenated with RGB channels. However, these methods fail to utilize the correlation between the two modalities. Meanwhile, the refinement stage of these methods is computationally expensive hence they cannot achieve real-time inference speed. Recently, \cite{wang2019densefusion} explored to fuse RGB and depth modalities at a late stage. It can achieve state-of-the-art performance while reaching almost real-time inference speed. Instead of direct feature concatenation as in \cite{wang2019densefusion}, we exploit the consistent and complementary information between two modalities by modeling the intra- and inter-modality correlation with a self-attention mechanism. Furthermore, we explore multiple fusion strategies to make the information flow within the framework more efficient. 

\subsection{Attention Mechanisms}
Attention mechanisms have been integrated in many deep learning-based computer vision and language processing tasks \cite{ye2021deep, ye2018hierarchical, eccv20ddag}, such as detection \cite{li2019ijcv}, classification \cite{wang2017cvpr} and visual question answering (VQA) \cite{gao2019dynamic}.There are many variants of the attention mechanisms, among which self-attention \cite{vaswani2017attention} has attracted lots of interests, due to its ability to model long-range dependencies while maintaining computational efficiency. Motivated by this work, we propose to integrate intra- and inter-modality correlation modelling with self-attention module for efficient fusion of RGB and depth information in 6D pose estimation. We explore multiple strategies for multi-model feature fusion. To our best knowledge, this is the first work to explore an efficient fusion of RGB and depth information in 6D object pose estimation with a self-attention mechanism. We show that our proposed method enables efficient exploitation of the context information from both RGB and depth modality and can achieve state-of-the-art accuracy in 6D pose estimation.

\begin{figure*}[htb]
	\centering
	\includegraphics[width=0.95\linewidth]{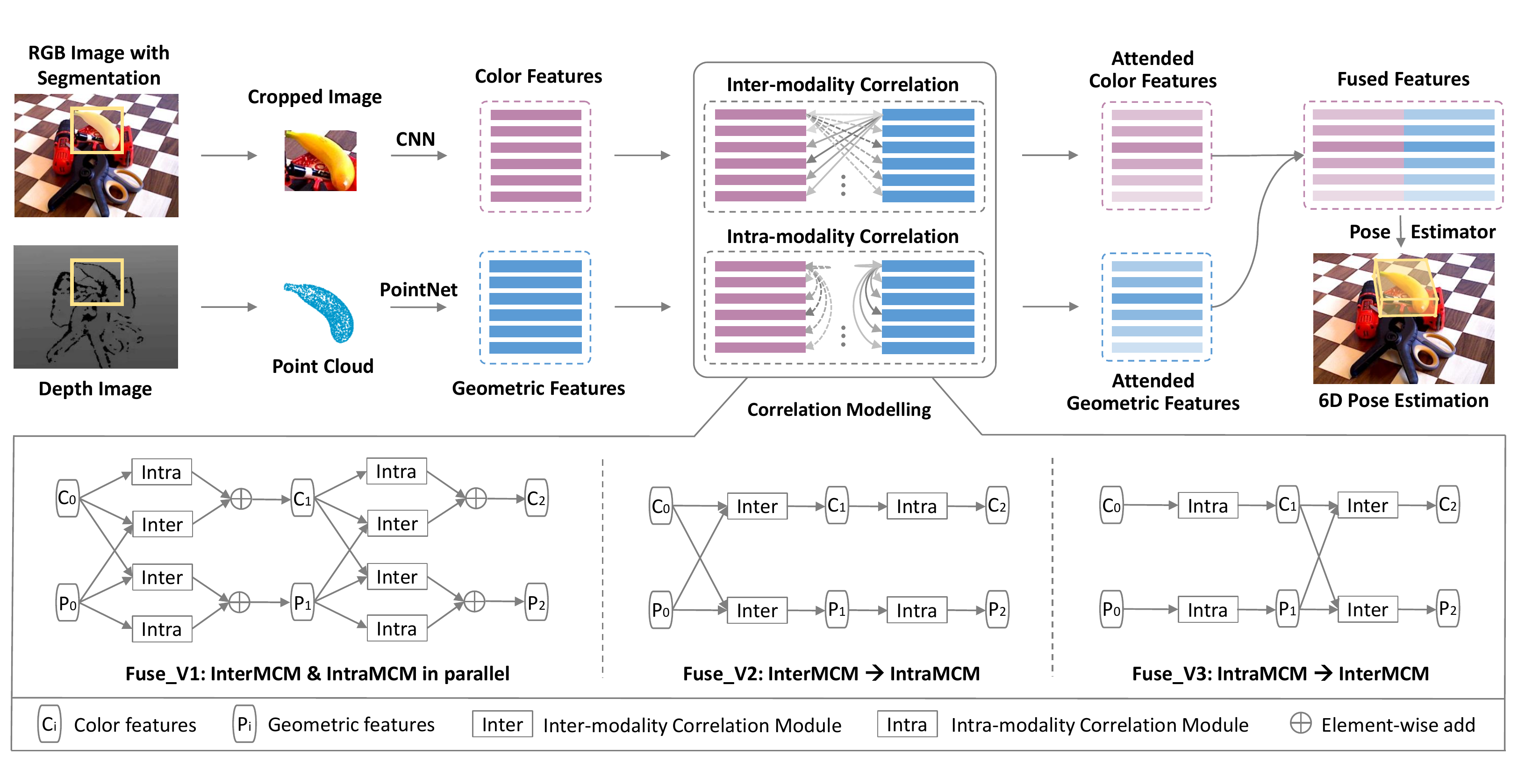}
	\caption{\textbf{Overview of Correlation Fusion (CF) framework for 6D pose estimation. } The Multi-modality Correlation Learning (MMCL) module contains Intra-modality Correlation  Modelling (IntraMCM) and Inter-modality Correlation  Modelling (InterMCM) modules. We also explore parallel (Fuse-V1) and sequential (Fuse-V2 and Fuse-V3) strategies to combine the two modules. This helps to efficiently model within and cross modalities dependency to capture the consistent and complementary information for accurate pose estimation.}
	\label{fig:overall}\vspace{-15pt}
\end{figure*}


\section{Methodology}

Given an RGB-D image and the 3D model of known objects, we aim to predict the 6D object pose $\mathbb{P}$ which is represented as a transformation $[R|T]$ in 3D space. 

Estimating 6D object pose from RGB image is challenging due to low textures, heavy occlusions and varying lighting conditions. Depth information 
provides extra geometric information to alleviate the problems. However, RGB and depth reside in two different modalities. Thus, efficient fusion schemes to maintain the modality-specific information as well as the complementary information from the other modalities are important for accurate pose estimation. 

Fig.~\ref{fig:overall} illustrates the proposed architecture. The first stage consists of semantic segmentation and feature extraction. The second stage is our main focus, which models the intra- and inter-correlation within and between RGB and depth modalities, followed by multiple module fusion strategies. 

Moreover, we have an additional stage which exploits an iterative refinement methodology to obtain final 6D pose estimation. We explain the detailed architecture in following subsections.

\subsection{Semantic Segmentation and Feature Extraction}
Firstly, we segment the target objects in the image with an existing semantic segmentation architecture proposed in \cite{xiang2017posecnn}. Specifically, given an image, the segmentation network generates a pixel-wise segmentation map to classify each image pixel into an object class. Then the RGB and depth images are cropped using the the bounding box of the predicted object. 

Secondly, the cropped RGB and depth images are processed separately to compute color and geometric features. For depth image $D$, the segmented depth pixels are first converted into $N$ 3D points with given camera intrinsics. Then, the points are fed to PointNet \cite{qi2017pointnet} variant (PNet) to obtain a geometric feature matrix $P \in \mathbb{R}^{{N}\times {d_{dep}}} $, where ${d_{dep}}$ refers to the dimension of PNet output feature vector for each point. Similarly, the cropped RGB image $I$ is applied through a CNN-based encoder-decoder architecture to produce a pixel-wise color feature matrix $C \in \mathbb{R}^{{N}\times {d_{rgb}}}$.


\subsection{Multi-modality Correlation Learning}
Our proposed Multi-modality Correlation Learning (MMCL) module contains Intra- and Inter-modality Correlation Modelling modules, where the former is to extract modality-specific features, while the latter is to extract modality-complement features.

\subsection{ (a) Intra-modality Correlation Modelling (IntraMCM)}
\begin{figure}[t]
	\begin{center}
		\includegraphics[width=1.0\linewidth]{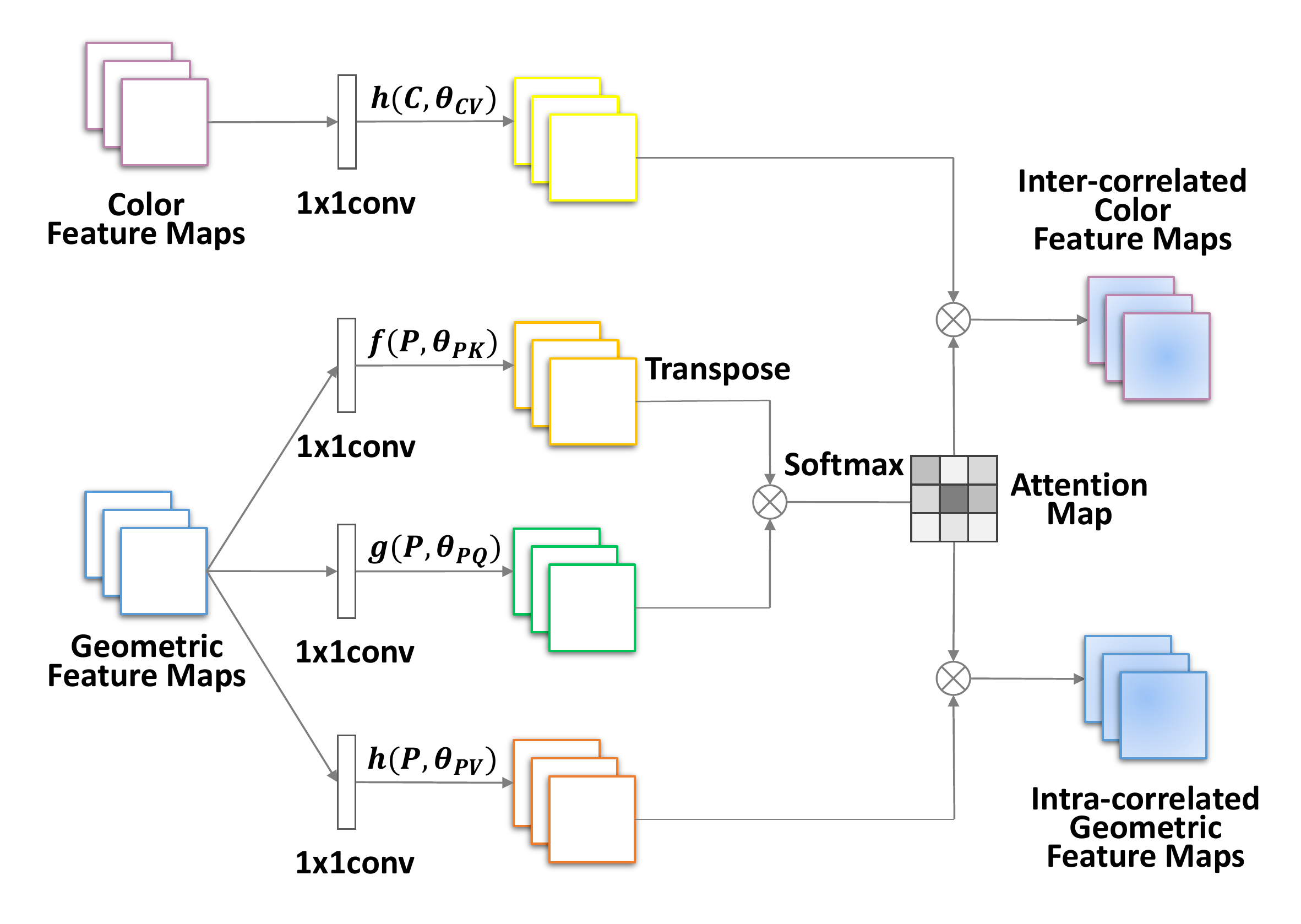}
	\end{center}
	\caption{Illustration of the proposed Intra- and Inter-modality Correlation Modelling modules. As these modules can be applied to color and geomtry features respectively in symmetric fashion, hence only the Intra-modality Correlation Modelling (IntraMCM) and Inter-modality Correlation Modelling (InterMCM) with geometric-centric features are shown. }
	\label{fig:MMCM}\vspace{-15pt}
\end{figure}

IntraMCM module is proposed to extract modality-specific discriminative features. The overall architecture of IntraMCM is illustrated in Fig. \ref{fig:MMCM}. Firstly, within each modality, features are transformed into query $Q$, key $K$ and value $V$ features with 1$\times$1 convolutions $\bm{f},\bm{g}$, and $\bm{h}$:

\begin{align}
&C_K = \bm{f}({C; \theta_{CK}}),  &P_K = \bm{f}({P; \theta_{PK}}), \\
&C_Q = \bm{g}({C; \theta_{CQ}}),  &P_Q = \bm{g}({P; \theta_{PQ}}), \\
&C_V = \bm{h}({C; \theta_{CV}}),  &P_V = \bm{h}({P; \theta_{PV}}), 
\label{eq:value}
\end{align}
where $C_K$, $C_Q$, $C_V \in \mathbb{R}^{ {N} \times d}$ are transformed color features, $P_K$, $P_Q$ and $P_V \in \mathbb{R}^{ {N} \times d}$ are transformed geometric features, $\theta$ are learned weight parameters and $d$ represents the common dimension of transformed features from both modalities. 

Then, the raw attention weights $C_A \in \mathbb{R}^{ {N} \times {N}}$ and $P_A  \in \mathbb{R}^{ {N} \times {N}}$ from RGB and depth modalities are obtained by computing the inner product $C_{Q} C_K^T$ and $P_{Q} P_K^T$ with row-wise softmax as follow:
\begin{align}
C_A = \textnormal{softmax}(C_Q C_K^T), \\
P_A = \textnormal{softmax}(P_Q P_K^T). 
\end{align}
Subsequently, $C_A$ and $P_A$ are used to weight information flow within RGB and depth modality, respectively as:
\begin{align}
C_{\bigtriangleup} = C_A \times C_V, \\
P_{\bigtriangleup} = P_A \times P_V,
\end{align}
where we denote the update of color and geometric feature maps as $C_{\bigtriangleup} \in \mathbb{R}^{N \times dim}$ and $P_{\bigtriangleup} \in \mathbb{R}^{N \times dim}$, respectively.

Then the element-wise addition is applied to combine $C_{\bigtriangleup}$ with original color feature maps $C$, which weights by two learnable parameter $\lambda_{CC}$ to obtain final feature maps $C_{\star}$ and this process is similarly applied to the geometric feature $P$ as:
\begin{align}
C_{\star} = C + \lambda_{CC} \times C_{\bigtriangleup}, \\
P_{\star} = P + \lambda_{PP} \times P_{\bigtriangleup}.
\end{align}
This facilitates the model to first learn from local information and gradually assign more weights for the non-local information. The $\lambda_{CC}$ and $\lambda_{PP}$ are initialized as 0 as in \cite{zhang2019self}. 

Therefore, the output of the IntraMCM module could capture both local and non-local color-to-color and geometry-to-geometry relations, and therefore maintain prominent modality-specific features for the subsequent pose estimation.

\subsection*{(b) Inter-Modality Correlation Modelling (InterMCM)}
The architecture of InterMCM is illustrated in Fig. \ref{fig:MMCM}. InterMCM is formulated to extract modality-complement features. The module first generates two attention maps $C_A \in \mathbb{R}^{ {N} \times {N}}$ and $P_A \in \mathbb{R}^{ {N} \times {N}}$ in the same way as IntraMCM module does. Then, we use the generated attention maps to weight features from the other modality and obtain the updated feature maps as $C_{\bigtriangledown} \in \mathbb{R}^{N \times d}$ and $P_{\bigtriangledown} \in \mathbb{R}^{N \times d}$:
\begin{align}
C_{\bigtriangledown} = P_A \times C_V, \\
P_{\bigtriangledown} = C_A \times P_V.
\end{align}
The color and geometric feature maps are further updated in the same way as in IntraMCM module,
\begin{align}
C_{\ast} = C + \lambda_{PC} \times C_{\bigtriangledown}, \\
P_{\ast} = P + \lambda_{CP} \times P_{\bigtriangledown}.
\end{align}
where $\lambda_{PC}$ and $\lambda_{CP}$ are learnable parameters initialized as 0, same as in IntraMCM module.

\subsection{Multi-modality Fusion Strategies}
\label{ssec::MMCF}
In our work, we explore multiple fusion strategies to effectively combine two information flows as illustrated in Fig. \ref{fig:overall} and the effectiveness of each updating scheme will be studied in Section \ref{ssec::experiment}. 

\textbf{Parallel update:} The IntraMCM and InterMCM modules are applied simultaneously, which is termed as \textbf{Fuse\_V1}.

\textbf{Sequential update:} The IntraMCM and InterMCM modules are applied sequentially. \textbf{Fuse\_V2} refers to first performs InterMCM and then IntraMCM; while \textbf{Fuse\_V3} refers to the opposite sequence.

\subsection{Pose Estimation and Refinement}

\textbf{Dense pose prediction.} After the color and geometric features are fused, we predict the object's 6D pose in a pixel-wise dense manner with a confidence score indicating how likely it is to be the ground true object pose. The dense prediction makes our method robust to occlusion and segmentation errors. During inference, the predicted pose with the highest confidence is selected as the final prediction.

\textbf{Iterative pose refinement.} We adopt a refiner network as in \cite{wang2019densefusion} for iterative pose refinement. We integrate the correlation modelling module into the pose refiner network in the same fashion as we applied in main network in Fig. \ref{fig:overall}. Specifically, at each iteration, we perform pixel-wise fusion of the original color features and the transformed geometric features with the predicted pose in the prediction, and then feed the fused pixel-wise features to pose refiner networks, which outputs the residual pose based on the predicted pose from the previous prediction. After $K$ iterations, the final pose estimation is obtained as:
\begin{align}
\hat{\mathbb{P}} = [R_K|T_K] \cdot [R_{K - 1}|T_{K - 1}] \cdot \dots \cdot [R_{0}|T_{0}].
\end{align}
In our implementation, we train refiner network after the main network converges. Such training process produces a descent performance with fast convergence speed.

\begin{table*}
	\begin{center}
		\centering
			\caption{The 6D pose estimation accuracy on \textbf{YCB-Video Dataset} in terms of the \textbf{ADD(-S) \textless{}2cm and the AUC of \textbf{ADD(-S)}}. The objects with bold name are considered as symmetric. All the methods use RGB-D images as input.\textbf{(Best zoom-in and view in pdf.)}}
	\label{exp:ycbauc}
		\small
		\scalebox{0.8}{
			\begin{tabular}{l|c c|c c|c c|c c|c c|c c|c c}
				\hline
				Methods & \multicolumn{2}{c}{PoseCNN\cite{xiang2017posecnn}} & \multicolumn{2}{c}{DenseFusion\cite{wang2019densefusion}} &  \multicolumn{2}{c}{OURS (IntraMCM)} & \multicolumn{2}{c}{OURS (InterMCM)} & \multicolumn{2}{c}{OURS (Fuse\_V1)} & \multicolumn{2}{c}{OURS (Fuse\_V2)} &\multicolumn{2}{c}{OURS (Fuse\_V3)}  \\ \hline
				Metrics & AUC       & \textless{}2cm      & AUC            & \textless{}2cm & AUC        & \textless{}2cm       & AUC          & \textless{}2cm         & AUC              & \textless{}2cm    & AUC              & \textless{}2cm  & AUC              & \textless{}2cm \\ \hline				
				002\_master\_chef\_can	           & 68.06 & 51.09      & 73.16 & 72.56            	& 87.61 & 88.37            	& 86.94 & 88.07           	& 86.18 & 86.28     & \textbf{92.47} & \textbf{98.71}           & 87.24 & 86.18  \\\hline
				003\_cracker\_box	               & 83.38 & 73.27           	& 94.21 	& 98.50            	& 94.80& \textbf{99.54}             	& 93.69 & 99.08            	& 91.79& 98.50      & \textbf{95.45} & 98.62           & 95.20& 99.19 \\\hline
				004\_sugar\_box	           & \textbf{97.15}   & 99.49         	& 96.50  & \textbf{100.00}          	& 93.73 & \textbf{100.00}           	& 95.06  & \textbf{100.00}           	& 95.68      & \textbf{100.00}       	& 96.69  & 99.92          	& 96.19 & 99.58 \\\hline
				005\_tomato\_soup\_can         	   & 81.77    	& 76.60        	& 85.42   & 82.99          	& 91.50  & 95.42          	& 90.23 & 93.19    & \textbf{92.73} & 95.56          	& 92.02  & \textbf{95.76}            	& 91.51& 95.56  \\\hline
				006\_mustard\_bottle  	   & \textbf{98.01}  & 98.60           	& 94.61 & 96.36            	& 92.27  & 98.04           	& 93.10& 98.60             	& 89.66& 91.04             	& 94.82  & 97.48          	& 95.29& \textbf{99.16}  \\ \hline                          	           	           	           	           	
				007\_tuna\_fish\_can	           & 83.87& 72.13             	& 81.88  & 62.28          	& 80.86& 69.69            	& 86.18 & 84.58           	& 85.94 & 83.45    & \textbf{88.85}  & 84.15          	& 85.27  & \textbf{86.31}\\\hline
				008\_pudding\_box	       & \textbf{96.62}  & \textbf{100.00}          	& 93.33 & 98.60            	& 91.69 & 97.13             	& 91.83  & 98.60           	& 91.76 & 99.07           	& 93.16 & 98.60            	& 94.10 & 98.13  \\\hline
				009\_gelatin\_box		   & \textbf{98.08}  & 100.00            	& 96.68  & 100.00           	& 95.35 & 100.00            	& 95.06  & 100.00           	& 95.92 & 100.00           	& 95.68 & 100.00             	& 97.28 & 100.00  \\\hline
				010\_potted\_meat\_can             & 83.47 & 77.94           	& 83.54 & 79.90           	& 85.01 & 83.55            	& 83.77& 80.81            	& 84.07	& 82.90     & \textbf{86.19} & 83.94           	& 86.03 & \textbf{84.07} \\\hline
				011\_banana	 	                   & 91.86 & 88.13           	& 83.49 & 88.13           	& 84.70& 81.79             	& 90.71& 98.68            	& 88.73 & 98.15    & \textbf{92.57} & \textbf{98.94}             	& 86.84 & 88.92 \\\hline                                    	            	           	            	     
				019\_pitcher\_base	 	   & \textbf{96.93} & 97.72           	& 96.78& 99.47            	& 95.76& 98.02            	& 96.55& \textbf{100.00}            	& 96.07 & \textbf{100.00}            	& 95.43& 98.42             	& 95.97 & 99.65 \\\hline
				021\_bleach\_cleanser	   & \textbf{92.54} & \textbf{92.71}            	& 89.93 & 90.96            	& 87.93 & 83.19           	& 89.10& 83.28            	& 90.19 & 89.70           	& 88.99	& 86.20            	& 89.00& 83.28  \\\hline                  	            	            	                        	 
				\textbf{024\_bowl}	               & 80.97 & 54.93      & \textbf{89.50} & 94.83           & 88.70 & \textbf{97.78}           	& 87.00 & 84.24           	& 86.32 & 90.64           	& 86.06& 94.33             	& 89.08 & 95.81  \\\hline
				025\_mug	                       & 81.08& 55.19             	& 88.92 & 89.62           	& 91.84& 92.77            	& 92.00 & 94.97           	& 91.06	& 91.98     & \textbf{93.51} & 94.81            & 93.44 & \textbf{96.38}  \\\hline
				035\_power\_drill	       & \textbf{97.66} & \textbf{99.24}           	& 92.55 	& 96.40           	& 92.05 & 95.65            	& 86.60 & 90.35            	& 85.05 	& 87.70           	& 82.89 	& 84.77           	& 93.52 & 98.20 \\\hline
				\textbf{036\_wood\_block}	       & 87.56 & 80.17      & \textbf{92.88} & \textbf{100.00}           & 91.44& 98.35            	& 90.16 & \textbf{100.00}           	& 91.46 & 99.59           	& 92.32 & 99.59           	& 92.35 & 98.76 \\\hline
				037\_scissors	                   & 78.36& 49.17            	& 77.89& 51.38     & \textbf{91.28} & 86.37           & 78.98& 67.40            	& 79.25& 64.70            	& 90.15 & \textbf{89.50}           	& 88.38& 86.74\\\hline  
				040\_large\_marker	               & 85.26 & 87.19            	& 92.95& \textbf{100.00}            	& 93.55& \textbf{100.00}            	& 93.84 & \textbf{100.00}           	& \textbf{94.10}& \textbf{100.00}    & 93.91  & 99.85          	& 93.82 & 99.85  \\\hline
				\textbf{051\_large\_clamp} & \textbf{75.19} & 74.86            	& 72.48  & \textbf{78.65}            	& 71.27& 78.51             	& 72.14& 77.95            	& 70.18 & 75.70           	& 70.31 & 76.69           	& 73.22& \textbf{78.65}  \\\hline
				\textbf{052\_extra\_large\_clamp}  & 64.38  & 48.83          	& 69.94 & 75.07           	& 70.11 & \textbf{76.83}    & \textbf{73.74}& 75.51            & 69.71 & 75.22           	& 69.53 & 74.49           	& 70.80 & 76.25 \\\hline
				\textbf{061\_foam\_brick}  & \textbf{97.23} 	& 100.00           	& 91.95  & 100.00           	& 94.36& 100.00            	& 94.15 & 100.00           	& 93.08 & 100.00            	& 94.62& 100.00          	& 94.89& 100.00  \\ \hline
				MEAN                               & 86.64 & 79.87             	& 87.55	& 88.37              	& 88.85& 91.48            	& 88.61& 91.21             	& 88.04 	& 90.96           	& 89.79& \textbf{93.08}      & \textbf{89.97}& 92.89  \\ \hline
			\end{tabular}
		} 
	\end{center}
	\vspace{-10pt}
\end{table*}


\section{Experiments}
\label{ssec::experiment}
\subsection{Datasets and Metrics}
We compare our method with the state-of-the-art methods on two commonly used datasets, namely YCB-Video~\cite{xiang2017posecnn} and LineMOD datasets~\cite{hinterstoisser2011iccv}. The pose estimation performance is evaluated by using average distance (ADD) metric \cite{hinterstoisser2012accv} and average  closest  point  distance (ADD-S) metric \cite{xiang2017posecnn}. The ADD metric is calculated by first transforming the model points with the predicted pose $[\hat{R}|\hat{T}]$ and the ground truth pose $[R|T]$, respectively, and then computing the mean of the pairwise distances between two sets of transformed points:
\begin{equation}
\textsc{ADD} = \frac{1}{m}\sum_{\mathbf{x} \in \mathcal{M}}\| (\mathbf{R} \mathbf{x} + \mathbf{T}) - (\mathbf{\tilde{R}} \mathbf{x} + \mathbf{\tilde{T}})  \|,
\end{equation}
where $\mathcal{M}$ denotes the 3D model points set and $m$ refers to number of points within the points set. The ADD-S metric \cite{xiang2017posecnn} is proposed for symmetric objects, where the matching between points sets is ambiguous for some views. ADD-S is defined as:
\begin{equation} \label{eq::dsym}
\textsc{ADD-S} = \frac{1}{m}\sum_{\mathbf{x}_1 \in \mathcal{M}} \min_{\mathbf{x}_2 \in \mathcal{M}} \| (\mathbf{R} \mathbf{x}_1 + \mathbf{T}) - (\mathbf{\tilde{R}} \mathbf{x}_2 + \mathbf{\tilde{T}})  \|.
\end{equation}

\subsection{Implementation Details}
\label{ssec::implementation}

We implement our method under the PyTorch \cite{paszke2017pytorch} framework. All the parameters except specified are initialized with PyTorch default initialization. Our model is trained using Adam optimizer \cite{kingma2014adam} with an initial learning rate at 1e-4. After the loss of estimator network falls under 0.016, then a decay of 0.3 is applied to further train the refiner network. The mini-batch size is set to $8$ for estimator network and $4$ for the refiner network.

\begin{table*}
	\begin{center}
		\centering
		\caption{The 6D pose estimation accuracy on the \textbf{LINEMOD Dataset} in terms of the \textbf{ADD(-S)} metric. The objects with bold name (glue and eggbox) are considered as symmetric. All the methods use RGB-D images as input.}
	\label{exp:linemod}
		\scalebox{1}{
			\begin{tabular}{l|c|c|c|c|c|c|c|c}
				\hline
				& SSD6D & BB8 & DenseFusion & OURS & OURS & OURS & OURS & OURS  \\
				&\cite{kehl2017ssd} &\cite{rad2017bb8} &\cite{wang2019densefusion} &(IntraMCM) &(InterMCM) &(Fuse\_V1) &(Fuse\_V2) &(Fuse\_V3) \\ \hline
				
				ape               & 65	& 40.4	& 92.3	& 94.9	& 95.2	& 94.8	& \textbf{95.6}	& 95.4    \\
				bench             & 80	& 91.8	& 93.2	& 93.7	& 94.0	& 96.1	& \textbf{96.9}	& 96.1    \\
				camera            & 78	& 55.7	& 94.4	& 97.5	& 95.6	& 96.0	& \textbf{97.9}	& 97.5    \\
				can               & 86	& 64.1	& 93.1	& 95.4	& 95.7	& 92.2	& \textbf{96.0}	& 95.0    \\
				cat               & 70	& 62.6	& 96.5	& 98.4	& 98.8	& \textbf{99.2}	& 97.8	& 99.1    \\ 
				driller           & 73	& 74.4	& 87.0	& 92.2	& 92.7	& 91.4	& \textbf{95.6}	& 94.7    \\
				duck              & 66	& 44.3	& 92.3	& \textbf{96.2}	& 95.1	& 95.7	& 95.7	& 95.8    \\
				\textbf{eggbox}   & \textbf{100}	& 57.8	& 99.8	& \textbf{100.0}	& 99.6	& \textbf{100.0}	& 99.9	& 99.9    \\
				\textbf{glue}     & \textbf{100}	& 41.2	& \textbf{100.0}	& 99.8	& 99.8	& 99.8	& 99.7	& 99.8    \\
				hole              & 49	& 67.2	& 92.1	& 95.2	& 95.6	& 95.8	& 96.7	& \textbf{97.1}    \\
				iron              & 78	& 84.7	& 97.0	& 95.8	& 96.2	& 97.4	& 97.8	& \textbf{98.4}    \\
				lamp              & 73	& 76.5	& 95.3	& 95.4	& 96.3	& 96.5	& \textbf{97.0}	& 96.8    \\
				phone             & 79	& 54.0	& 92.8	& 97.3	& \textbf{97.5}	& 95.6	& 97.0	& 97.4    \\ \hline
				MEAN              & 77	& 62.7	& 94.3	& 96.3	& 96.3	& 96.2	& \textbf{97.2}	& 97.1    \\ \hline
			\end{tabular}\vspace{-10pt}
		} 
	\end{center}

\end{table*}

\subsection{Experiment Analysis}
\label{exp:analysis}
\subsubsection{Ablation Study}



In this section, we perform an ablation study on the effectiveness of each component, i.e., IntraMCM and InterMCM, in the proposed network. We consider five ablation models including \textbf{Intra-Only}: our method with only IntraMCM component; \textbf{Inter-Only}: our method with only InterMCM component; \textbf{Fuse\_V1}: our method with parallel update on both IntraMCM and InterMCM simultaneously; \textbf{Fuse\_V2}: our method which first performs IntraMCM and then InterMCM; \textbf{Fuse\_V3}: our method which first performs InterMCM and then IntraMCM. Table \ref{exp:ycbauc} and Table \ref{exp:linemod} records the experimental results of the ablation models on YCB-VIDEO and LINEMOD datasets, respectively. It is observed that Intra-Only, Inter-Only and Fuse\_V1 produce comparable pose estimation performance. Both Fuse\_V2 and Fuse\_V3 significantly outperform Intra-Only, Inter-Only and Fuse\_V1 models. From these observations, we conclude that the sequential updating (Fuse\_V2 and Fuse\_V3) generally outperforms parallel updating (Fuse\_V1). Since Fuse\_V2 and Fuse\_V3 perform similarly well and therefore we also conclude that the fusion effect is not sensitive to the specific updating order of IntraMCM and InterMCM.

\begin{figure*}[htb]
	\centering
	\includegraphics[width=1\linewidth]{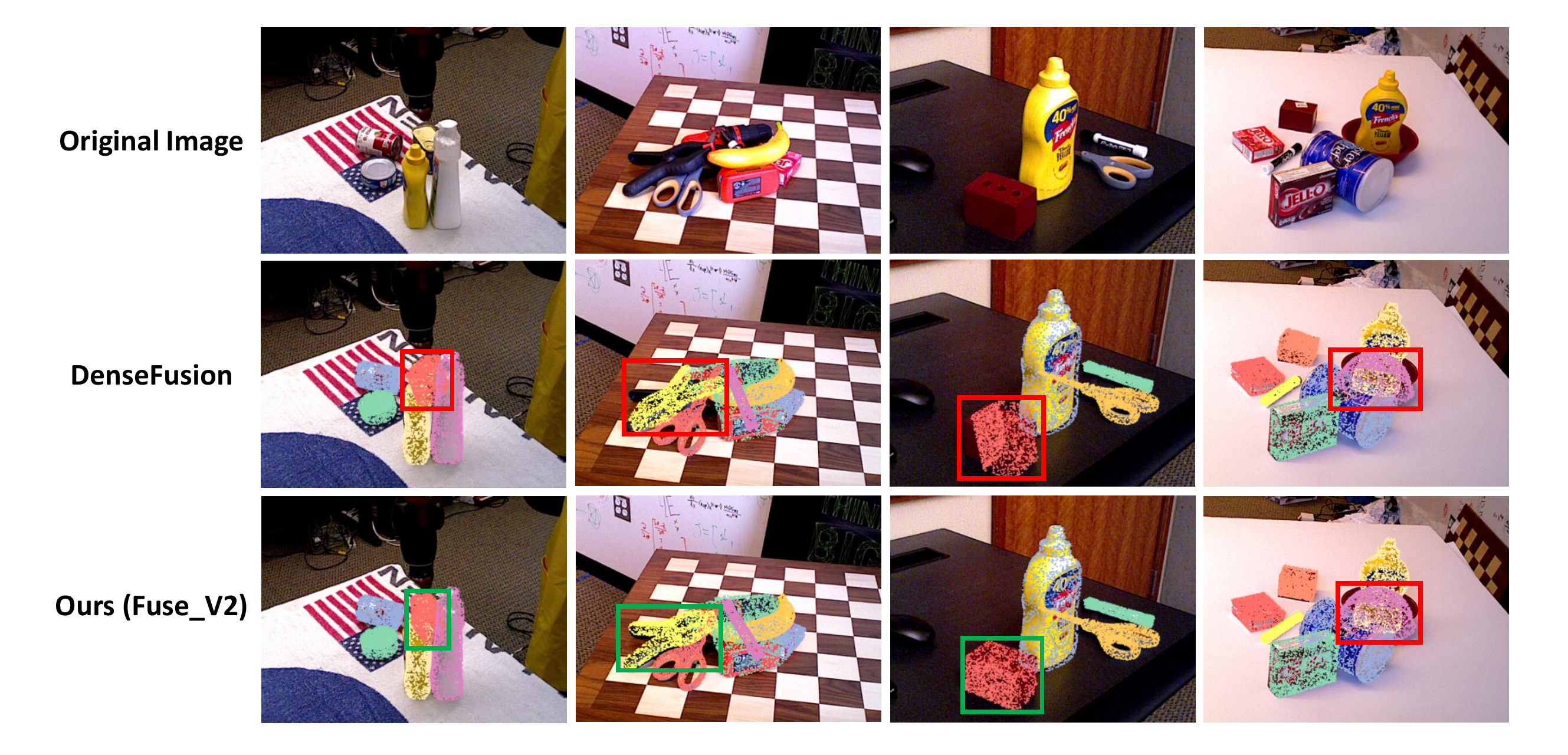}
	\caption{\textbf{Visualizations of results on the YCB-Video Dataset.} The first row is original RGB image, the second row is from DenseFusion, and third row is our proposed method Fuse\_V2. The red boxes show the cases with poor pose estimation while the green boxes shows the ones with good pose estimation.}
	\label{fig:visualization}
\end{figure*}

\subsubsection{Comparison with the State-of-the-Arts}
We also compare our method with the state-of-the-art methods which take RGB-D images as input and output 6D object poses on YCB-Video and LineMOD dataset. 

\textbf{Comparisons on YCB-Video dataset.} The results in terms of ADD(-S) AUC and ADD(-S) \textless{}2cm metrics are presented in Table \ref{exp:ycbauc}. For both metrics, our method is superior to the state-of-the-art methods PoseCNN \cite{xiang2017posecnn} and DenseFusion \cite{wang2019densefusion}. Specifically, our method with Fuse\_V2 outperforms PoseCNN by a margin of 13.21\% and DenseFusion by 4.71\% in terms of the ADD(-S) \textless{}2cm metric. 

\textbf{Comparisons on LineMOD dataset.} Table \ref{exp:linemod} summarizes the comparison with \cite{kehl2017ssd, rad2017bb8, wang2019densefusion} in terms of ADD(-S) metric on LineMOD dataset. For benchmarking methods, SSD-6D \cite{kehl2017ssd} and BB8 \cite{rad2017bb8} obtain the initial 6D pose estimation with RGB image as input and then use depth image for pose refinement.  DenseFusion \cite{wang2019densefusion} takes both RGB and depth images for pose estimation and pose refinement. Comparing with these methods, our proposed method achieves the best performance by taking advantage of the correlation information from RGB and depth modalities.  


\subsection{Efficiency and Qualitative Results}

On average, the running time of our full model 6D pose estimation for one image is 49.8ms, including 23.6ms for the semantic segmentation forward propagation, 17.3ms for pose estimation forward propagation, and 8.9ms for the forward propagation of refiner on a single NVIDIA GTX 2080ti GPU. Our method can run in real-time on GPU at around 20fps.

In Fig. \ref{fig:visualization}, we present some qualitative results on the examples from YCB-Video dataset, for both DenseFusion \cite{wang2019densefusion} and our proposed method. From the figure, compared with DenseFusion, it is shown that our method could achieve better pose estimation either under heavy occlusions (shown by the meat can in the first column) or for the predictions on symmetric objects (shown by the large clamp and foam brick in the second and the third columns, respectively). However, for the extremely challenging cases like predicting a symmetric object under heavy occlusions (shown by the bowl appeared in the fourth row images), both our method and DenseFusion generate less accurate estimation. More visualization results are presented in the attached video.


\subsection{Robotic Grasping Experiments}
We also carry out robotic grasping experiments in both simulation and real world to demonstrate that our method is effective in assisting robots to correctly grasp objects by providing accurate object pose estimation. We have provided a video \footnote{\url{https://www.dropbox.com/s/ht2lqew2raogiu3/ICPR2020_video.mp4?dl=0}.} demonstrating a robot grasping task in both simulation and real-world scenarios. 


\textbf{Grasping in simulation.} We compare the proposed method with DenseFusion \cite{wang2019densefusion} in Gazebo simulation environment. We retrain both models with data collected from the environment. We place four objects from YCB-Video dataset in five random locations and four random orientations on the table. The robot arm aligns gripper with the predicted object pose to grasp the target object. The robot arm makes 20 attempts to grasp each object, with 80 grasps in total for each comparing method. The results are shown in Table \ref{exp:graspexp}. Thanks to the correlation fusion framework, our method has a significantly higher pick-up success rate than that in \cite{wang2019densefusion}. Some visualization results are presented in the attached video.


\begin{table}
	\begin{center}
		\caption{Success rate for the grasping experiments with robotic arm in simulation environment of Gazebo. }
	\label{exp:graspexp}
		\centering
		\small
		\scalebox{0.7}{
			\begin{tabular}{|l|c|c|c|c|}
				\hline
				Success Attempts (\%)     & tomato\_soup\_can      & mustard\_bottle     & banana      & bleach\_cleanser  \\ \hline	
			    DenseFusion \cite{wang2019densefusion}	& 80.0    & 70.0   & 55.0    & 65.0        \\\hline
				Ours	                                & 90.0    & 85.0   & 75.0 	& 80.0   	 \\\hline
			\end{tabular}
		} 
	\end{center}
	\vspace{-15pt}
\end{table}

\textbf{Grasping in real world.} We also apply our method in a real-world robot grasping task, where the robot arm is used to pick up objects from a table. Without further fine-tuning on real testing data, our model can predict sufficiently accurate object pose for the grasping task. Some visualization results are presented in the attached video.

\section{Conclusion}

In this paper, we have proposed a novel Correlation Fusion framework with intra- and inter-modality correlation learning for 6D object pose estimation. The IntraMCM module is designed to learn prominent modality-specific features and the InterMCM module is to capture complement modality features. Subsequently, multiple fusion schemes are explored to further improve the performance on 6D pose estimation. Extensive experiments conducted on YCB, LINEMOD dataset and a real-world robot grasping task demonstrate the superior performance of our method to several benchmarking methods.

\section{Acknowledgement}
This research is supported by the Agency for Science, Technology and Research (A*STAR) under its AME Programmatic Funding Scheme (Project A18A2b0046).

\bibliography{bib/main}

\begin{thebibliography}{10}
\providecommand{\url}[1]{#1}
\csname url@samestyle\endcsname
\providecommand{\newblock}{\relax}
\providecommand{\bibinfo}[2]{#2}
\providecommand{\BIBentrySTDinterwordspacing}{\spaceskip=0pt\relax}
\providecommand{\BIBentryALTinterwordstretchfactor}{4}
\providecommand{\BIBentryALTinterwordspacing}{\spaceskip=\fontdimen2\font plus
\BIBentryALTinterwordstretchfactor\fontdimen3\font minus
  \fontdimen4\font\relax}
\providecommand{\BIBforeignlanguage}[2]{{%
\expandafter\ifx\csname l@#1\endcsname\relax
\typeout{** WARNING: IEEEtran.bst: No hyphenation pattern has been}%
\typeout{** loaded for the language `#1'. Using the pattern for}%
\typeout{** the default language instead.}%
\else
\language=\csname l@#1\endcsname
\fi
#2}}
\providecommand{\BIBdecl}{\relax}
\BIBdecl

\bibitem{LoghmaniPCV19}
M.~R. Loghmani, M.~Planamente, B.~Caputo, and M.~Vincze, ``Recurrent
  convolutional fusion for {RGB-D} object recognition,'' \emph{{IEEE} Robotics
  and Automation Letters}, vol.~4, no.~3, pp. 2878--2885, 2019.

\bibitem{MousavianAFK17}
A.~Mousavian, D.~Anguelov, J.~Flynn, and J.~Kosecka, ``3{D} bounding box
  estimation using deep learning and geometry,'' in \emph{CVPR}, 2017, pp.
  5632--5640.

\bibitem{tremblay2018corl}
J.~Tremblay, T.~To, B.~Sundaralingam, Y.~Xiang, D.~Fox, and S.~Birchfield,
  ``Deep object pose estimation for semantic robotic grasping of household
  objects,'' in \emph{Conference on Robot Learning}, 2018, pp. 306--316.

\bibitem{li2018fast}
M.~Li and K.~Hashimoto, ``Fast and robust pose estimation algorithm for bin
  picking using point pair feature,'' in \emph{2018 24th International
  Conference on Pattern Recognition (ICPR)}.\hskip 1em plus 0.5em minus
  0.4em\relax IEEE, 2018, pp. 1604--1609.

\bibitem{xiang2017posecnn}
Y.~Xiang, T.~Schmidt, V.~Narayanan, and D.~Fox, ``Posecnn: A convolutional
  neural network for 6d object pose estimation in cluttered scenes,''
  \emph{arXiv preprint arXiv:1711.00199}, 2017.

\bibitem{li2018deepim}
Y.~Li, G.~Wang, X.~Ji, Y.~Xiang, and D.~Fox, ``Deepim: Deep iterative matching
  for 6d pose estimation,'' in \emph{Proceedings of the European Conference on
  Computer Vision (ECCV)}, 2018, pp. 683--698.

\bibitem{manhardt2018eccv}
F.~Manhardt, W.~Kehl, N.~Navab, and F.~Tombari, ``Deep model-based 6d pose
  refinement in rgb,'' in \emph{Proceedings of the European Conference on
  Computer Vision (ECCV)}, 2018, pp. 800--815.

\bibitem{pavlakos2017icra}
G.~Pavlakos, X.~Zhou, A.~Chan, K.~G. Derpanis, and K.~Daniilidis, ``6-dof
  object pose from semantic keypoints,'' in \emph{2017 IEEE International
  Conference on Robotics and Automation (ICRA)}.\hskip 1em plus 0.5em minus
  0.4em\relax IEEE, 2017, pp. 2011--2018.

\bibitem{oberweger2018eccv}
M.~Oberweger, M.~Rad, and V.~Lepetit, ``Making deep heatmaps robust to partial
  occlusions for 3d object pose estimation,'' in \emph{Proceedings of the
  European Conference on Computer Vision (ECCV)}, 2018, pp. 119--134.

\bibitem{peng2019pvnet}
S.~Peng, Y.~Liu, Q.~Huang, X.~Zhou, and H.~Bao, ``Pvnet: Pixel-wise voting
  network for 6dof pose estimation,'' in \emph{Proceedings of the IEEE
  Conference on Computer Vision and Pattern Recognition}, 2019, pp. 4561--4570.

\bibitem{hu2019cvpr}
Y.~Hu, J.~Hugonot, P.~Fua, and M.~Salzmann, ``Segmentation-driven 6d object
  pose estimation,'' in \emph{Proceedings of the IEEE Conference on Computer
  Vision and Pattern Recognition}, 2019, pp. 3385--3394.

\bibitem{tekin2018cvpr}
B.~Tekin, S.~N. Sinha, and P.~Fua, ``Real-time seamless single shot 6d object
  pose prediction,'' in \emph{Proceedings of the IEEE Conference on Computer
  Vision and Pattern Recognition}, 2018, pp. 292--301.

\bibitem{rad2017bb8}
M.~Rad and V.~Lepetit, ``Bb8: a scalable, accurate, robust to partial occlusion
  method for predicting the 3d poses of challenging objects without using
  depth,'' in \emph{Proceedings of the IEEE International Conference on
  Computer Vision}, 2017, pp. 3828--3836.

\bibitem{zhu2016cvpr}
H.~Zhu, J.-B. Weibel, and S.~Lu, ``Discriminative multi-modal feature fusion
  for rgbd indoor scene recognition,'' in \emph{Proceedings of the IEEE
  Conference on Computer Vision and Pattern Recognition}, 2016, pp. 2969--2976.

\bibitem{jafari2018ipose}
O.~H. Jafari, S.~K. Mustikovela, K.~Pertsch, E.~Brachmann, and C.~Rother,
  ``i{P}ose: instance-aware 6d pose estimation of partly occluded objects,'' in
  \emph{Asian Conference on Computer Vision}.\hskip 1em plus 0.5em minus
  0.4em\relax Springer, 2018, pp. 477--492.

\bibitem{kehl2017ssd}
W.~Kehl, F.~Manhardt, F.~Tombari, S.~Ilic, and N.~Navab, ``{SSD-6D}: Making
  {RGB}-based {3D} detection and {6D} pose estimation great again,'' in
  \emph{Proceedings of the IEEE International Conference on Computer Vision},
  2017, pp. 1521--1529.

\bibitem{brachmann2014eccv}
E.~Brachmann, A.~Krull, F.~Michel, S.~Gumhold, J.~Shotton, and C.~Rother,
  ``Learning 6d object pose estimation using 3d object coordinates,'' in
  \emph{European conference on computer vision}.\hskip 1em plus 0.5em minus
  0.4em\relax Springer, 2014, pp. 536--551.

\bibitem{krull2015iccv}
A.~Krull, E.~Brachmann, F.~Michel, M.~Ying~Yang, S.~Gumhold, and C.~Rother,
  ``Learning analysis-by-synthesis for 6d pose estimation in rgb-d images,'' in
  \emph{Proceedings of the IEEE International Conference on Computer Vision},
  2015, pp. 954--962.

\bibitem{michel2017iccv}
F.~Michel, A.~Kirillov, E.~Brachmann, A.~Krull, S.~Gumhold, B.~Savchynskyy, and
  C.~Rother, ``Global hypothesis generation for 6d object pose estimation,'' in
  \emph{Proceedings of the IEEE Conference on Computer Vision and Pattern
  Recognition}, 2017, pp. 462--471.

\bibitem{zeng2017icra}
A.~Zeng, K.-T. Yu, S.~Song, D.~Suo, E.~Walker, A.~Rodriguez, and J.~Xiao,
  ``Multi-view self-supervised deep learning for 6d pose estimation in the
  amazon picking challenge,'' in \emph{2017 IEEE International Conference on
  Robotics and Automation (ICRA)}.\hskip 1em plus 0.5em minus 0.4em\relax IEEE,
  2017, pp. 1386--1383.

\bibitem{wang2019densefusion}
C.~Wang, D.~Xu, Y.~Zhu, R.~Mart{\'\i}n-Mart{\'\i}n, C.~Lu, L.~Fei-Fei, and
  S.~Savarese, ``Densefusion: 6d object pose estimation by iterative dense
  fusion,'' in \emph{Proceedings of the IEEE/CVF Conference on Computer Vision
  and Pattern Recognition}, 2019, pp. 3343--3352.

\bibitem{hinterstoisser2011iccv}
S.~Hinterstoisser, S.~Holzer, C.~Cagniart, S.~Ilic, K.~Konolige, N.~Navab, and
  V.~Lepetit, ``Multimodal templates for real-time detection of texture-less
  objects in heavily cluttered scenes,'' in \emph{2011 international conference
  on computer vision}.\hskip 1em plus 0.5em minus 0.4em\relax IEEE, 2011, pp.
  858--865.

\bibitem{ye2021deep}
M.~Ye, J.~Shen, G.~Lin, T.~Xiang, L.~Shao, and S.~C. Hoi, ``Deep learning for
  person re-identification: A survey and outlook,'' \emph{IEEE Transactions on
  Pattern Analysis and Machine Intelligence}, 2021.

\bibitem{ye2018hierarchical}
M.~Ye, X.~Lan, J.~Li, and P.~Yuen, ``Hierarchical discriminative learning for
  visible thermal person re-identification,'' in \emph{Proceedings of the AAAI
  Conference on Artificial Intelligence}, vol.~32, no.~1, 2018.

\bibitem{eccv20ddag}
M.~Ye, J.~Shen, D.~J. Crandall, L.~Shao, and J.~Luo, ``Dynamic dual-attentive
  aggregation learning for visible-infrared person re-identification,'' in
  \emph{European Conference on Computer Vision (ECCV)}, 2020.

\bibitem{li2019ijcv}
H.~Li, Y.~Liu, W.~Ouyang, and X.~Wang, ``Zoom out-and-in network with map
  attention decision for region proposal and object detection,''
  \emph{International Journal of Computer Vision}, vol. 127, no.~3, pp.
  225--238, 2019.

\bibitem{wang2017cvpr}
F.~Wang, M.~Jiang, C.~Qian, S.~Yang, C.~Li, H.~Zhang, X.~Wang, and X.~Tang,
  ``Residual attention network for image classification,'' in \emph{Proceedings
  of the IEEE Conference on Computer Vision and Pattern Recognition}, 2017, pp.
  3156--3164.

\bibitem{gao2019dynamic}
P.~Gao, Z.~Jiang, H.~You, P.~Lu, S.~C. Hoi, X.~Wang, and H.~Li, ``Dynamic
  fusion with intra-and inter-modality attention flow for visual question
  answering,'' in \emph{Proceedings of the IEEE Conference on Computer Vision
  and Pattern Recognition}, 2019, pp. 6639--6648.

\bibitem{vaswani2017attention}
A.~Vaswani, N.~Shazeer, N.~Parmar, J.~Uszkoreit, L.~Jones, A.~N. Gomez,
  {\L}.~Kaiser, and I.~Polosukhin, ``Attention is all you need,'' in
  \emph{Advances in neural information processing systems}, 2017, pp.
  5998--6008.

\bibitem{qi2017pointnet}
C.~R. Qi, H.~Su, K.~Mo, and L.~J. Guibas, ``Pointnet: Deep learning on point
  sets for 3d classification and segmentation,'' in \emph{Proceedings of the
  IEEE Conference on Computer Vision and Pattern Recognition}, 2017, pp.
  652--660.

\bibitem{zhang2019self}
H.~Zhang, I.~Goodfellow, D.~Metaxas, and A.~Odena, ``Self-attention generative
  adversarial networks,'' in \emph{International Conference on Machine
  Learning}.\hskip 1em plus 0.5em minus 0.4em\relax PMLR, 2019, pp. 7354--7363.

\bibitem{hinterstoisser2012accv}
S.~Hinterstoisser, V.~Lepetit, S.~Ilic, S.~Holzer, G.~Bradski, K.~Konolige, and
  N.~Navab, ``Model based training, detection and pose estimation of
  texture-less 3d objects in heavily cluttered scenes,'' in \emph{Asian
  conference on computer vision}.\hskip 1em plus 0.5em minus 0.4em\relax
  Springer, 2012, pp. 548--562.

\bibitem{paszke2017pytorch}
A.~Paszke, S.~Gross, F.~Massa, A.~Lerer, J.~Bradbury, G.~Chanan, T.~Killeen,
  Z.~Lin, N.~Gimelshein, L.~Antiga \emph{et~al.}, ``Pytorch: An imperative
  style, high-performance deep learning library,'' \emph{arXiv preprint
  arXiv:1912.01703}, 2019.

\bibitem{kingma2014adam}
D.~P. Kingma and J.~Ba, ``Adam: A method for stochastic optimization,''
  \emph{arXiv preprint arXiv:1412.6980}, 2014.

\end{thebibliography}
\bibliographystyle{sty/IEEEtran}

\end{document}